%% file: ok_transformer.tex
\pdfoutput=1

\documentclass[11pt]{article}

\usepackage{acl}

\usepackage{times}
\usepackage{latexsym}

\usepackage[T1]{fontenc}

\usepackage[utf8]{inputenc}

\usepackage{microtype}
\usepackage{mathtools}
\usepackage{subcaption}
\usepackage{amssymb}
\usepackage{mathtools}
\usepackage{booktabs}
\newcommand{\loss}[0]{{\mathcal{L}}}
\newcommand{\nop}[1]{}

\title{Enhancing Natural Language Representation with Large-Scale Out-of-Domain Commonsense}

\author{
  Wanyun Cui, Xingran Chen \\
  Shanghai University of Finance and Economics \\
  \texttt{cui.wanyun@sufe.edu.cn, xingran.chen.sufe@gmail.com} \\
}

\begin{document}
\maketitle

\begin{abstract}
We study how to enhance text representation via textual commonsense. We point out that commonsense has the nature of domain discrepancy. Namely, commonsense has different data formats and is domain-independent from the downstream task. This nature brings challenges to introducing commonsense in general text understanding tasks. A typical method of introducing textual knowledge is continuing pre-training over the commonsense corpus. However, it will cause catastrophic forgetting to the downstream task due to the domain discrepancy. In addition, previous methods of directly using textual descriptions as extra input information cannot apply to large-scale commonsense.

In this paper, we propose to use large-scale out-of-domain commonsense to enhance text representation. In order to effectively incorporate the commonsense, we proposed OK-Transformer (\underline{O}ut-of-domain \underline{K}nowledge enhanced \underline{Transformer}). OK-Transformer effectively integrates commonsense descriptions and enhances them to the target text representation. In addition, OK-Transformer can adapt to the Transformer-based language models (e.g. BERT, RoBERTa) for free, without pre-training on large-scale unsupervised corpora. We have verified the effectiveness of OK-Transformer in multiple applications such as commonsense reasoning, general text classification, and low-resource commonsense settings. \footnote{The code is available in \url{https://github.com/chenxran/ok-transformer}}
\end{abstract}

\input{ok_intro}

\input{ok_pdef}

\input{ok_method}
\input{ok_exp}

\input{ok_conclusion}

\section*{Acknowledgments and Disclosure of Funding}
We thank Wenting Ba for her valuable plotting assistance. This paper was supported by National Natural Science Foundation of China (No. 61906116), by Shanghai Sailing Program (No. 19YF1414700).

\bibliography{ok_casual}
 \bibliographystyle{acl_natbib}
 
\input{ok_append}

\end{document}

%% file: ok_intro.tex
\section{Introduction}
\label{sec:intro}
\begin{figure*}[!htb]
	\centering
		\includegraphics[scale=.5]{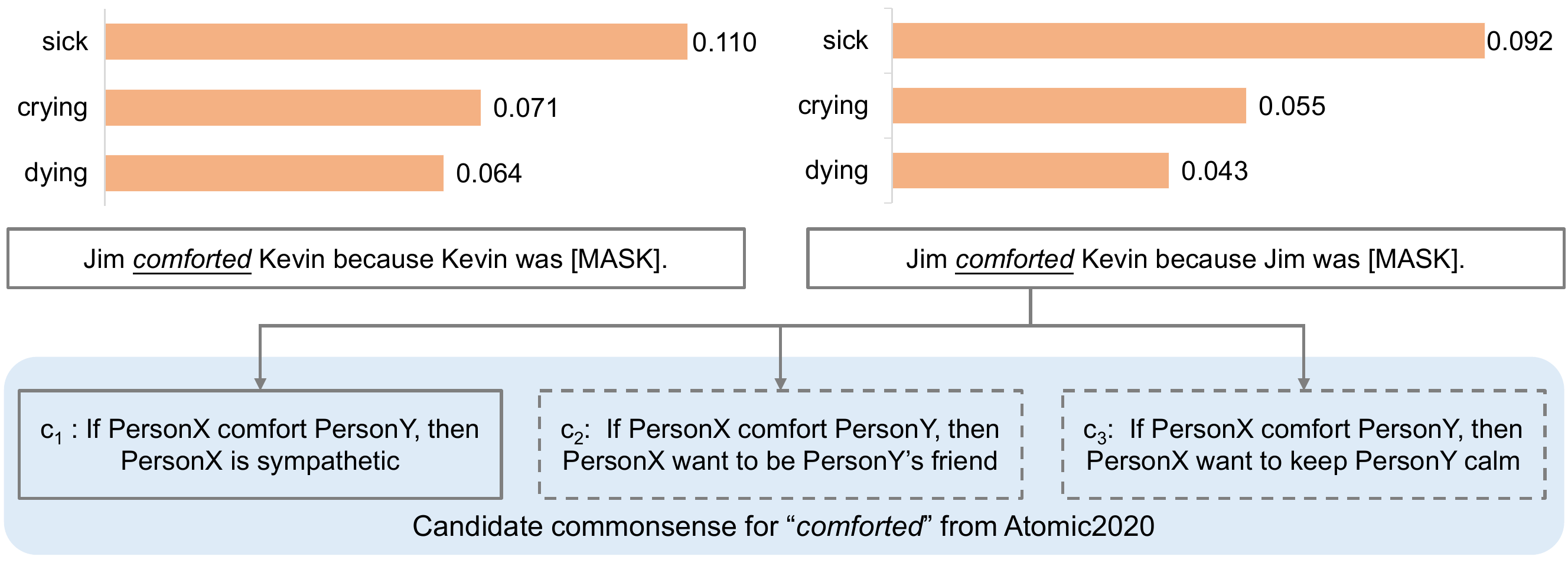}
\caption{The prediction of [MASK] by BERT. BERT cannot distinguish between {\it Jim} and {\it Kevin} in {\it Jim comforted Kevin because}. }
\label{fig:commonsense_list}
\vspace{-0.3cm}
\end{figure*}

Although unsupervised language models have achieved big success on many tasks~\cite{devlin2018bert}, they are incapable of learning low-frequency knowledge. For example, in the masked language model task in Fig.~\ref{fig:commonsense_list}, even if we replace ``Kevin was'' (left) with ``Jim was'' (right), BERT~\cite{devlin2018bert} still predicts the masked word as sick, crying, dying, etc. This is because similar texts in its training corpus rarely describe the subject of ``comforted''.
To improve the model's ability to generalize and understand low-frequency knowledge, we propose to incorporate commonsense into language models. In Fig.~\ref{fig:commonsense_list}, to make correct predictions, we need to enhance the language model with the commonsense $c_1$.

However, commonsense has the nature of {\it domain discrepancy}. The downstream task and the commonsense knowledge have distribution discrepancies. Taking the commonsense knowledge base we use (i.e. ATOMIC2020~\cite{hwang2020comet}) as an example, the distribution discrepancy is specifically manifested in (1) their data formats. The format of a commonsense description usually belongs to some specific patterns (e.g. “... As a result ...”, “... Because ...”), while the downstream tasks can have arbitrary patterns. (2) The commonsense belongs to the domain of event causality, while the downstream tasks may belong to arbitrary domains.



Here we highlight the challenges caused by the domain discrepancy. To introduce external textual knowledge to a pre-trained language model, a common practice is to continue pre-training the language model on the corpus of the external knowledge~\cite{gururangan2020don,sun2019finetune}. However, the study~\cite{gururangan2020don} also found that continuing pre-training requires external knowledge and downstream tasks to have similar domains. Due to its domain discrepancy, introducing commonsense through continuing pre-training will cause catastrophic forgetting to downstream tasks, thereby injuring the effectiveness. We have verified this empirically in Sec~\ref{sec:exp:introduction}. Therefore, the domain discrepancy prevents us from introducing commonsense by continuing pre-training.

To enhance the representation of the target text with external commonsense, we propose to directly use its candidate commonsense as an extra input. Our setup is different from a typical natural language understanding setup since the latter one only takes the target text as the input~\cite{devlin2018bert}. We argue that our setup -- where the commonsense is introduced explicitly as input -- is a more practicable setup to introduce out-of-domain commonsense that cannot be learned through pre-training. As far as we know, ExpBERT~\cite{murty2020expbert} is the closest setup to us. It also uses external knowledge (manually constructed templates) as the input.



Another challenge is the {\bf scale} of the commonsense. Although ExpBERT also allows extra textual commonsense as input, it only captures small-scale commonsense with a fixed size. In addition, when we introduce commonsense from a large-scale knowledge base for general purpose (i.e. ATOMIC2020), unrelated commonsense (e.g. $c_2$ and $c_3$ in Fig.~\ref{fig:commonsense_list}) will certainly occur. However, ExpBERT lacks the ability to distinguish related and unrelated commonsense. Therefore, the power of large-scale commonsense knowledge was restricted in ExpBERT. We will verify this empirically in Sec~\ref{sec:exp:introduction}.

In order to incorporate the large-scale out-of-domain commonsense, we propose the OK-Transformer (\underline{O}ut-of-domain \underline{K}nowledge enhanced \underline{Transformer}) on the basis of Transformer~\cite{vaswani2017attention}. OK-Transformer has two modules. The knowledge enhancement module is used to encode the target text with commonsense, and the knowledge integration module is used to encode and integrate all candidate commonsense. OK-Transformer has two advantages. First, it fully represents the contextual information of the textual commonsense. Second, it can be adapted to existing pre-trained language models (e.g. BERT and RoBERTa) for free.
That is, we are able to adapt OK-Transformer to the pre-trained language models, without pre-training OK-Transformer over large-scale unsupervised corpora from scratch.

Some other methods are related to our work, such as introducing {\it structured} knowledge~\cite{peters2019knowledge,zhang2019ernie,guan2020knowledge,zhou2018commonsense} and {\it plain text} knowledge~\cite{guu2020realm} in language models. These methods do not represent the specific inductive bias of commonsense knowledge and therefore are not suitable to introduce commonsense. We will compare these studies with more details in Sec~\ref{sec:related}.


%% file: ok_pdef.tex

\section{Related work}
\label{sec:related}
In this section, we compare different ways to introduce knowledge into language models. We divide the knowledge introduction methods into (1) continuing pre-training method~\cite{gururangan2020don,sun2019finetune} and (2) explicit introduction in the downstream task~\cite{guu2020realm,murty2020expbert}. 

{\bf Continuing pre-training} the language model is effective when the external knowledge is similar to the downstream task~\cite{gururangan2020don,sun2019finetune}. However, commonsense and downstream tasks have domain discrepancies, so continuing pre-training is unsuitable for introducing commonsense. We have empirically verified this in Sec~\ref{sec:exp:introduction}.

{\bf Introducing explicit knowledge in downstream tasks} We classify the knowledge into structured knowledge, plain text, and semi-structured knowledge, depending on its form. The entries of {\bf structured knowledge} are represented as individual embeddings~\cite{peters2019knowledge,zhang2019ernie,guan2020knowledge,zhou2018commonsense}, while commonsense descriptions in this paper can be represented more accurately by the contextual information of their word sequences. 



\section{Problem Setup: Commonsense as the Extra Input}

We consider a text classification task where the text $x$ and its label $y$ are provided for training. Assuming that the candidate commonsense descriptions for enhancing $x$ come from a large-scale commonsense knowledge base (i.e. ATOMIC2020), we retrieve candidate commonsense for $x$ as the extra input. We denote the commonsense descriptions for $x$ as $cs(x)=\{c_1 \cdots c_n\}$, where each $c_i$ is a commonsense description. The retrieval process will be shown in Sec~\ref{sec:exp}. The model takes both $x$ and $cs(x)$ as the input. Since ATOMIC2020 contains if-then knowledge for general purposes, the problem setup can be expanded to a broad range of text understanding tasks. The goal of training is to find parameter $\theta$ that minimizes the loss of training examples given the texts and candidate commonsense descriptions:
\begin{equation}
\small
\label{eqn:goal}
\mathrm{{\arg\min}_{\theta} \mathbb{E}_{(x,y)\in train} \loss(f(x,cs(x)|\theta),y)}
\end{equation}
where $\mathrm{f(\cdot|\theta)}$ is the model taking $x$ and $cs(x)$ as inputs, $\loss$ is the loss function.

%% file: ok_method.tex
\section{OK-Transformer}

In this section, we propose OK-Transformer based on Transformer to introduce extra commonsense descriptions. We first show OK-Transformer on an abstract level in Sec~\ref{sec:framework}. Then we elaborate two modules within it, i.e. knowledge enhancement and knowledge integration, in Sec~\ref{sec:knowledge:input} and Sec~\ref{sec:method:output}, respectively.

\subsection{Framework}
\label{sec:framework}
\begin{figure}[!tb]
	\centering
		\includegraphics[scale=.45]{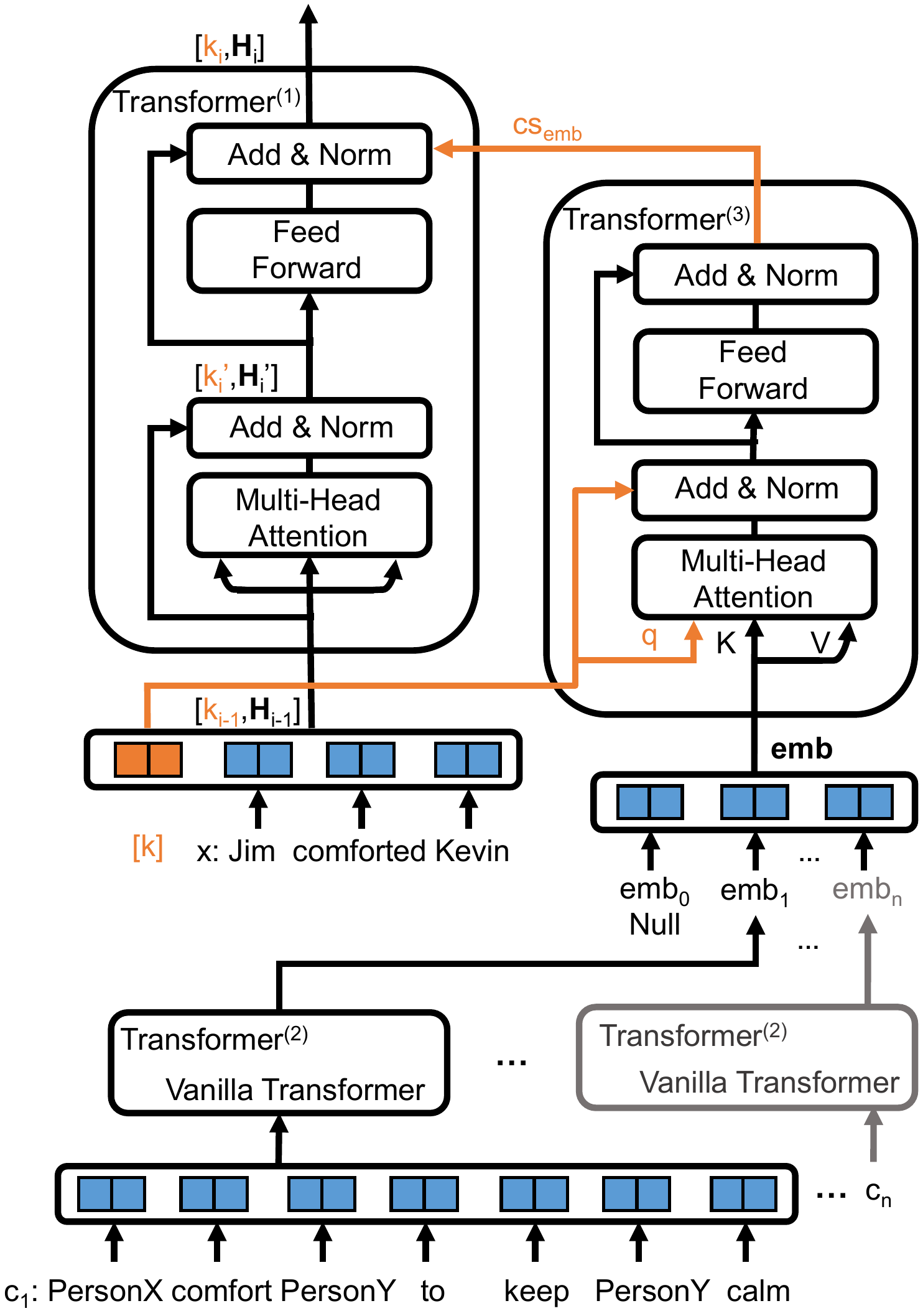}
\caption{OK-Transformer. $\mathrm{Transformer^{(1)}}$ encodes the target text $x$ with enhanced commonsense $k_i$. $\mathrm{Transformer^{(2)}}$ encodes each individual commonsense description. $\mathrm{Transformer^{(3)}}$ integrates all candidate commonsense descriptions and transfers knowledge to $\mathrm{Transformer^{(1)}}$.}
\label{fig:ok_transformer}
\end{figure}

In this subsection, we show how our OK-Transformer works at an abstract level. For the target sentence $x$, OK-Transformer takes both $x$ and $cs(x)$ as inputs. To incorporate all the information of $x$ and $cs(x)$, the OK-Transformer contains three vanilla Transformers, denoted by $\mathrm{Transformer^{(1)(2)(3)}}$. The knowledge enhancement module uses $\mathrm{Transformer^{(1)}}$ to encode the target text. Compared with the vanilla Transformer, $\mathrm{Transformer^{(1)}}$ leverages a new knowledge token to represent the commonsense that interacts with other words. The knowledge integration module encodes each individual commonsense description by $\mathrm{Transformer^{(2)}}$, and then integrates all candidate commonsense descriptions by $\mathrm{Transformer^{(3)}}$. This is shown in Fig.~\ref{fig:ok_transformer}.

\subsection{Knowledge Enhancement Module}
\label{sec:knowledge:input}
The knowledge enhancement module allows commonsense knowledge to enhance the representation of the target text.

{\bf Interaction between words and commonsense.} We use $\mathrm{Transformer^{(1)}}$ to represent the interaction between words of the target text $x$. In addition, we introduce a special token $[k]$ to represent the commonsense knowledge. We denote it as the knowledge token. $\mathrm{Transformer^{(1)}}$ encodes all words and the knowledge token together via multi-head attention. Formally, given word sequence $\mathrm{x=w_1, \cdots, w_n}$, $\mathrm{Transformer^{(1)}}$ accepts a sequence of $n+1$ word-piece tokens: $\mathrm{[k], \; w_1, \cdots w_n}$. We denote the knowledge embedding and word embeddings produced by the $i$-th layer of $\mathrm{Transformer^{(1)}}$ as $k_i \in \mathbb{R}^{d}$ and $\bf{H}_i \in \mathbb{R}^{n \times d}$, respectively. The $\mathrm{Transformer^{(1)}}$ block first uses a multi-head self-attention layer followed by a residual connection and a layer normalization to model their interactions:
\begin{equation}
\small
\label{eqn:kh}
\begin{aligned}
    & \mathrm{k_i',{\bf H}_i'= LayerNorm( [k_{i-1},{\bf H}_{i-1}] +} \\ 
    & \mathrm{MultiHeadAttn([k_{i-1},{\bf H}_{i-1}],[k_{i-1},{\bf H}_{i-1}],[k_{i-1},{\bf H}_{i-1}]))} \\
\end{aligned}
\end{equation}
where $\mathrm{[k_{i-1},{\bf H}_{i-1}] \in \mathbb{R}^{(n+1) \times d}}$ means appending $k_{i-1}$ at the front of ${\bf H}_{i-1}$. $\mathrm{[k_{i-1},{\bf H}_{i-1}]}$ is used as the query, key, and value in the multi-head attention.

{\bf Knowledge update} The vanilla Transformer projects $\mathrm{k_i', \;{\bf H}_i'}$ in Eq.~\eqref{eqn:kh} to the output space with a multi-layer perceptron neural network (MLP). Compared to the vanilla Transformer, we use an extra update operation to update the knowledge token by the integrated commonsense knowledge after the MLP. As in the vanilla Transformer, the update layer is followed by a residual connection and a layer normalization. This can be formulated by:
\begin{equation}
\small
\begin{aligned}
&\mathrm{k_{i}=LayerNorm(k_i'+MLP(k_i')+ cs_{emb})} \\
&\mathrm{{\bf H}_{i}=LayerNorm({\bf H}_{i}'+MLP({\bf H}_i'))}
\end{aligned}
\end{equation}
where $cs_{emb}$ is the embedding of the commonsense computed by the knowledge integration module in Sec~\ref{sec:method:output}. 

\subsection{Knowledge Integration Module}
\label{sec:method:output}
The knowledge integration module encodes all candidate commonsense descriptions and integrates them. We first use $\mathrm{Transformer^{(2)}}$ to represent each candidate commonsense description. Then, we use $\mathrm{Transformer^{(3)}}$ to integrate all candidate commonsense, and transfer the integrated knowledge to the knowledge enhancement module.

{\bf Representing single commonsense} We use a vanilla Transformer as $\mathrm{Transformer^{(2)}}$ to model each candidate commonsense description. For all the retrieved commonsense $cs(x)=\{c_1,\cdots,c_n\}$, we compute the embedding $emb_j$ of each commonsense description $c_j$ by:
\begin{equation}
\small
\mathrm{emb_j = Transformer^{(2)}(c_j)}
\end{equation}

{\bf Knowledge integration} We integrate all candidate commonsense by $\mathrm{Transformer^{(3)}}$. Since not all the candidate commonsense leads to high confidence prediction as we have discussed in Sec~\ref{sec:intro}, we need to select relevant commonsense and ignore irrelevant commonsense. Transformer is adequate to conduct this selection. Specifically, in the query-key-value mechanism in Transformer, we use the embedding of the knowledge token in $\mathrm{Transformer^{(1)}}$ as the query of $\mathrm{Transformer^{(3)}}$. 
and the commonsense embeddings by $\mathrm{Transformer^{(2)}}$ as keys and values of $\mathrm{Transformer^{(3)}}$. Then, we integrate representations of all different commonsense descriptions based on their similarities with the knowledge token.

$\mathrm{Transformer^{(3)}}$ also uses multi-head attention to allow the knowledge token to interact with the candidate commonsense in multiple ways. The output of multi-head self-attention is followed by a residual connection and a layer normalization.
\begin{equation}
\small
\begin{aligned}
\mathrm{cs_{emb}}= & \mathrm{LayerNorm(k_{i-1} }\\ 
& + \mathrm{MultiHeadAttn(k_{i-1},{\bf emb},{\bf emb}))}
\end{aligned}
\end{equation}
where $\mathrm{{\bf emb}=[emb_1,\cdots,emb_n]}$ denotes the sequence of embeddings of all candidate commonsense descriptions. We then apply a residual connection and a layer normalization to it.

{\bf Null Commonsense} Some target texts may not have valid commonsense from ATOMIC2020 to enhance their representations. Therefore, we refer to the settings of REALM~\cite{guu2020realm} to add a null commonsense into the candidate commonsense of all target texts. We denote the null commonsense as $c_0$. Matching to the null commonsense indicates that the commonsense knowledge base cannot help enhance the target text.

\section{Adaptation to Pre-trained Language Models}
In this section, we take BERT as an example to illustrate how we adapt OK-Transfomer to existing pre-trained language models. We denote the adapted model as OK-BERT. An important manifestation of the effectiveness of the  Transformer structure is its applications in large-scale pre-trained models (e.g. BERT, RoBERTa). In order to introduce external knowledge, many other studies conduct training over large-scale unsupervised corpus~\cite{peters2019knowledge,xiong2019pretrained}. However, OK-Transformer is able to directly adapt to the existing pre-trained language models for free. In other words, when adapting OK-Transformer to OK-BERT, we directly use the parameters of each Transformer layer of BERT to initialize the OK-Transformer layers of OK-BERT. This property greatly improves the applicability of OK-BERT. In the rest of this section, we will describe how $\mathrm{Transformer^{(1)}}$, $\mathrm{Transformer^{(2)}}$, and $\mathrm{Transformer^{(3)}}$ are adapted respectively in Sec~\ref{sec:layer_adapt}, and how to fine-tune OK-BERT in Sec~\ref{sec:train}.

\subsection{Layer-by-Layer Adaptation}
\label{sec:layer_adapt}
\begin{figure}[!htb]
	\centering
		\includegraphics[scale=.45]{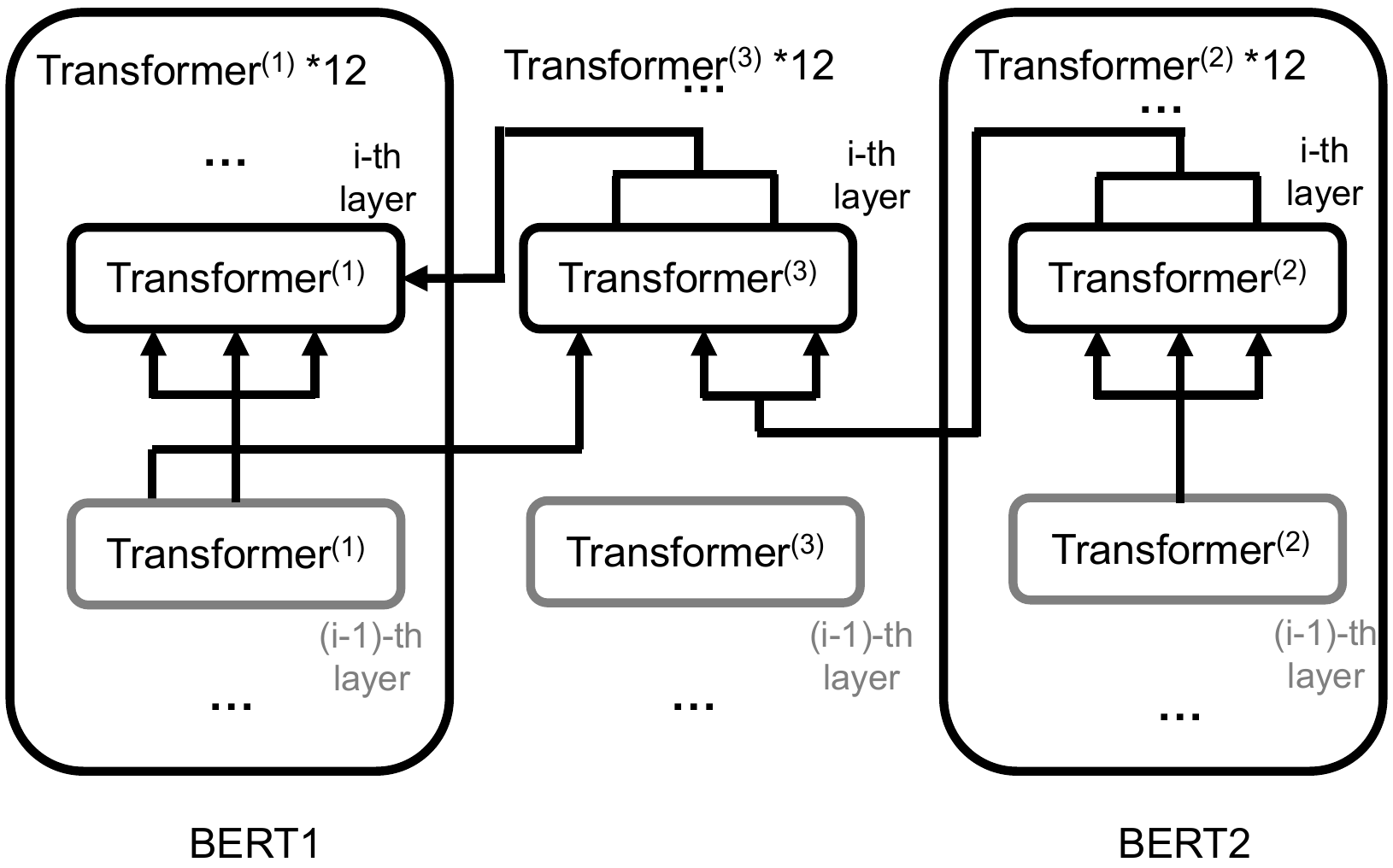}
\caption{The architecture of OK-BERT. We only draw edges that connect to the $i$-th layer.}
\label{fig:ok_bert}
\end{figure}

The OK-BERT we designed uses two original BERTs to serve as $\mathrm{Transformer^{(1)}}$ and $\mathrm{Transformer^{(2)}}$, respectively. We denote them as BERT1 and BERT2.
We connect the $\mathrm{Transformer^{(1)}}$ and $\mathrm{Transformer^{(2)}}$ in the corresponding layer of each BERT by $\mathrm{Transformer^{(3)}}$. Therefore, OK-BERT makes full use of the multi-layer structure of BERT, while allowing commonsense in the knowledge token to fully interact with the target text in each layer. The architecture is shown in Fig.~\ref{fig:ok_bert}.

$\mathrm{\bf Transformer^{(1)}}$ We adapt the Transformer of BERT1 to $\mathrm{Transformer^{(1)}}$ in the knowledge enhancement module of OK-Transformer. Note that the original BERT's tokens are $\mathrm{[CLS]\; w_1 \cdots w_L\; [SEP]}$ (for a single sentence) or $\mathrm{[CLS] \; w_1 \cdots w_{m}\; [SEP]\; w_{m+1} \cdots w_{L}\; [SEP]}$ (for a sentence pair). We follow~\cite{wang2020cross} and use a special token $[k]$ as the knowledge token. 
When tokenizing sentences, we insert the $[k]$ token after the $[CLS]$ token for each given text. In this way, the input tokens become $\mathrm{[CLS]\; [k]\; w_1 \;\cdots w_L\; [SEP]}$ or $\mathrm{[CLS]\; [k]\; w_1 \cdots w_m \; [SEP]\; w_{m+1} \cdots w_{L}\; [SEP]}$
, respectively. This simple modification allows us to use $[k]$ as the knowledge token in the knowledge enhancement module.


$\mathrm{\bf Transformer^{(2)}}$ We adapt each Transformer layer of BERT2 to the $\mathrm{Transformer^{(2)}}$ layer. The adaptation is straightforward since $\mathrm{Transformer^{(2)}}$ uses the vanilla Transformer structure. We use the encoding of the $[CLS]$ token in each corresponding layer as the commonsense representation $emb_j$ to enhance the representation of the corresponding layer in BERT1.

$\mathrm{\bf Transformer^{(3)}}$ For each pair of corresponding $\mathrm{Transformer^{(1)}}$ and $\mathrm{Transformer^{(2)}}$ from the same layer, we use one $\mathrm{Transformer^{(3)}}$ to connect them to transfer the information from BERT2 to BERT1.

In summary, when adapting to BERT-base with 12 Transformer layers, OK-BERT contains 12 $\mathrm{Transformer^{(1)}}$ layers for BERT1, 12 $\mathrm{Transformer^{(2)}}$ layers for BERT2, and 12 $\mathrm{Transformer^{(3)}}$ layers for layer-wise knowledge integration.

\subsection{Parameter Initialization and Model Training}
\label{sec:train}
In our implementation, BERT1 and BERT2 have independent parameters. We use the parameters of BERT to initialize both BERT1 and BERT2. The parameters of $\mathrm{Transformer^{(3)}}$ layers are randomly initialized. For downstream tasks, we then fine-tune all the parameters in the fashion of end2end. 

%% file: ok_exp.tex
\section{Experiments}
\label{sec:exp}

We evaluate the effectiveness of our proposed models in three scenarios: cloze-style commonsense reasoning, text classification, and low-resource commonsense settings. All the experiments run over a computer with 4
Nvidia Tesla V100 GPUs.

{\bf Models} We consider adapting OK-Transformer to BERT and RoBERTa, which are denoted as OK-BERT and OK-RoBERTa, respectively. We use the BERT-base and RoBERTa-large from the HuggingFace Transformer library~\cite{wolf2020transformers}. 

{\bf Implementation details for candidate knowledge retrieval} For a given text $x$, we retrieve candidate commonsense from ATOMIC2020. We use the if-then descriptions in ATOMIC2020 (e.g. Fig.~\ref{fig:commonsense_list}). Since these descriptions cover 173k different verb phrases -- one of the fundamental elements of language -- the retrieval is applicable to a broad range of downstream text understanding tasks.

We use a simple retrieval method. We simply consider word segments with window size 5 of the input text $x$. All the commonsense descriptions matching one of these text segments will be regarded as the candidate commonsense descriptions $c_i \in cs(x)$.


\subsection{Commonsense Reasoning}

\subsubsection{Setup}

{\bf Datasets} We consider the following commonsense reasoning benchmarks: WSC273~\cite{levesque2012winograd}, PDP~\cite{morgenstern2016planning}, Winogender~\cite{rudinger2018gender}, WinoGrande~\cite{sakaguchi2019winogrande}, CommonsenseQA~\cite{talmor2019commonsenseqa} and PhysicalQA~\cite{bisk2020piqa}.


{\bf Model details} 
Due to the different implementations between \cite{kocijan19acl} and~\cite{sakaguchi2019winogrande}, in this paper, we also follow their settings to compare with them, respectively. For~\cite{kocijan19acl}, we conduct disambiguation tasks directly through masked language modeling in OK-BERT. For the latter one, we convert cloze-style problems to multiple-choice classification problems in OK-RoBERTa. In particular, we replace the target pronoun of one query sentence with each candidate reference, then put the new sentences into the language model. We use a single linear layer and a softmax layer over the encoding of its $[CLS]$ token to compute the probability of each new sentence, and select the one with the highest probability as the pronoun disambiguation result.

{\bf Hyperparameters of pre-training} We follow~\cite{kocijan19acl,sakaguchi2019winogrande} to first pre-train models for 30 and 3 epochs over WSCR~\cite{kocijan19acl} or WinoGrande~\cite{sakaguchi2019winogrande}, respectively. Then we fine-tune models over specific tasks. We use AdamW as the optimizer with learning rate 5e-6, which is selected from $\{2e-5, 1e-5,5e-6\}$. We set the batch size to 8.

\begin{table}[!htb]
\small
\setlength{\tabcolsep}{2pt}
\centering
\begin{tabular}{l c c}
\toprule
Model & WSC & PDP \\ \hline
KEE\cite{liu2016commonsense} & 52.8 & 58.3 \\
WKH~\cite{emami2018generalized} & 57.1 & -  \\
MAS~\cite{klein2019attention}  & 60.3  & 68.3  \\
DSSM~\cite{wang2019unsupervised} & 63.0 & 75.0  \\
LM\cite{trinh2018simple} & 63.8 & 70.0   \\
CSS~\cite{klein2020contrastive} & 69.6 & 90.0   \\
GPT2~\cite{radford2019language} & 70.7 & -    \\
BERT-large+WSCR~\cite{kocijan19acl} & 71.4 & 79.2 \\
HNN~\cite{he2019hybrid} & 75.1 & 90.0  \\ 
Human~\cite{sakaguchi2019winogrande}  & 96.5 & 92.5\\ \hline
BERT+WSCR  & 66.3 & 85.0 \\
{\bf OK-BERT+WSCR }         & {\bf 67.4} & {\bf 86.7 }  \\ \hline
RoB.+WinoGrande  & 90.1 & 87.5 \\
{\bf OK-RoB.+WinoGrande } & {\bf 91.6} & {\bf 91.7} \\
\bottomrule
\end{tabular}
\caption{Results on WSC and PDP. RoB. denotes RoBERTa.}
\label{tab:main1}
\end{table}

\begin{table}[!htb]
\centering
\small
\setlength{\tabcolsep}{2pt}
\setlength{\tabcolsep}{1.3pt}
\begin{tabular}{lcc}
    \toprule
    Model & WinoGen. & WinoGran. \\ \hline
    WikiCREM~\cite{kocijan2019wikicrem}                & 82.1 & -   \\
    WinoGrande~\cite{sakaguchi2019winogrande} & 94.6 & 79.3 \\ \hline
    BERT+WSCR                      & 68.2 & 51.4 \\
    {\bf OK-BERT+WSCR}                   & {\bf 72.4} & {\bf 53.4} \\ \hline
    RoB.+WinoGrande            & 94.6 & 79.3 \\
    {\bf OK-RoB.+WinoGrande }        & {\bf 96.2} & {\bf 79.6} \\
    \bottomrule
\end{tabular}
\caption{Results on WinoGender and WinoGrande.}
\label{tab:main2}
\end{table}

\begin{table}[!htb]
\centering
\setlength{\tabcolsep}{2.2pt}
\small
\begin{tabular}{  l | c c}
\toprule
Model & CommonsenseQA & PhysicalQA \\ \hline
BERT               & 55.86 & 68.71   \\
{\bf OK-BERT}                   & {\bf 56.27} & {\bf 69.09} \\ \hline
RoBERTa           & 73.55 & 79.76 \\
{\bf OK-RoBERTa }        & {\bf 75.92} & {\bf 80.09} \\
\bottomrule
\end{tabular}
\caption{Results on CommonsenseQA and PhysicalQA.}
\label{tab:main3}
\end{table}

\subsubsection{Results}
We compare our models with state-of-the-art commonsense reasoning models in Table~\ref{tab:main1},~\ref{tab:main2}, and~\ref{tab:main3}. It can be seen that our models outperform other models in most settings. This verifies the effectiveness of our proposed models for commonsense reasoning.

{\bf Ablations} In Table~\ref{tab:main1},~\ref{tab:main2}, and~\ref{tab:main3} we also compare OK-BERT with BERT. We found that OK-BERT with OK-Transformers effectively improved the accuracy of BERT with Transformers. Similar results can be found between OK-RoBERTa and RoBERTa. This shows that the proposed OK-Transformer improves pre-trained language models by adapting to them for free, i.e. without retraining on large-scale unsupervised corpora.

\subsection{General Text Classification}
\label{sec:exp:glue}
We use MRPC, CoLA, RTE, STS-B, SST-2, and QNLI in the GLUE dataset~\cite{wang2018glue} to verify the effectiveness of the proposed models on general text classification tasks. We did not evaluate over MNLI, because our model needs to represent the corresponding $n$ commonsense for each sentence, which is too costly for MNLI. We believe that this efficiency problem can be solved by further applying model compression~\cite{iandola2020squeezebert}, but this is beyond the scope of this paper. It can be seen from Table~\ref{tab:glue} that OK-BERT and OK-RoBERTa outperform their baselines. 

\begin{table*}[!htb]
\centering
{
    \begin{tabular}{lcccccc}
    \toprule
    GLUE Task       & MRPC & CoLA & RTE & QNLI & STS-B & SST-2 \\ \hline
    BERT       & 86.27/90.21 & \textbf{59.50} & 71.43 & 91.20 & 89.35/88.93 & 91.97 \\
    OK-BERT         & \textbf{87.25/90.84} & 58.29 & \textbf{73.65} & \textbf{91.58} & \textbf{89.82/89.46} & \textbf{93.69} \\ \hline
    RoBERTa   & 90.44/93.15 & 66.57 & 84.11 & 94.00 & 91.83/91.95 & 95.70 \\
    OK-RoBERTa      & \textbf{91.91/94.24} & \textbf{66.89} & \textbf{86.28} & \textbf{94.41} & \textbf{92.41/92.20} & \textbf{96.10} \\
    \bottomrule
    \end{tabular}
}
\caption{Results on text classification tasks. Models are evaluated by the dev split from GLUE. }
\label{tab:glue}
\end{table*}

\subsection{Commonsense Introduction Methods}
\label{sec:exp:introduction}
{\bf Continue pre-train} In the introduction section, we mentioned that a typical method of introducing textual knowledge is continuing pre-training~\cite{gururangan2020don,sun2019finetune}. However, due to the domain discrepancy of commonsense, this method will cause catastrophic forgetting. To verify this intuition, in this subsection we compare with the continuing pre-trained model. We first continue pre-training the language model over ATOMIC2020, then fine-tune it over the target task.


{\bf ExpBERT}~\cite{murty2020expbert}
We also compare our OK-Transformer with ExpBERT, another model that is able to introduce textual knowledge. In Sec~\ref{sec:intro}, we mentioned that ExpBERT is not applicable to large-scale commonsense knowledge bases for its disability to select related commonsense and ignore unrelated commonsense. To verify this, we use the retrieved candidate commonsense descriptions from ATOMIC2020 as the additional explanations for ExpBERT. ExpBERT concatenates all the embedding of a fixed number of commonsense, which is inflexible for ATOMIC2020. For this reason, we fix the number of commonsense to 48. If there are more than 48 candidate commonsense descriptions for one sample, we will randomly select 48 of them. Otherwise, we will pad null commonsense to it. In our experiments, we also apply ExpBERT to RoBERTa~\cite{liu2019roberta} (i.e. ExpRoBERTa).

\begin{table*}[!htb]
\centering
\begin{tabular}{lccccccc}
\toprule
                 & MRPC        & CoLA  & RTE   & QNLI  & STS-B       & SST-2 & WSC273 \\ \hline
BERT        & 86.27/90.21 & \textbf{59.50} & 71.43 & 91.20 & 89.35/88.93 & 91.97    & 66.30  \\
BERT-continue & 83.58/88.81         & 54.70  & 62.09 & 90.24 & 87.41/87.46 & 91.74 & 63.00  \\
ExpBERT         & 85.78/89.79 & 58.29 & 62.82 & 87.06 & 84.78/84.67 & 91.51 & --\\ 
{\bf OK-BERT}    & \textbf{87.25/90.84} & 58.29 & \textbf{73.65} & \textbf{91.58} & \textbf{89.82/89.46} & \textbf{93.69}  & \textbf{67.40} \\
\hline
RoBERTa    & 90.44/93.15 & 66.57 & 84.11 & 94.00 & 91.83/91.95 & 95.70  & 90.10  \\
RoBERTa-continue  & 87.01/90.38 & 61.74 & 74.01 & 93.61 & 89.57/89.66 & 95.99 & 87.91 \\
ExpRoBERTa      & 89.46/92.22 & \textbf{66.90}  & 83.39 & 93.78 & 89.81/89.94 & 95.99 & -- \\
{\bf OK-RoBERTa} & \textbf{91.91/94.24} & 66.89 & \textbf{86.28} & \textbf{94.41} & \textbf{92.41/92.20} & \textbf{96.10} & \textbf{91.58} \\
\bottomrule
\end{tabular}
\caption{Comparison of different commonsense introduction approaches. Continuing pre-training even injures the effectiveness. On the other hand, using OK-Transformers to introduce external knowledge achieves better results than using Transformer.}
\label{tab:continue}
\end{table*}

We show the results in Table~\ref{tab:continue}. We do not report the results of ExpBERT on WSC273, as ExpBERT cannot solve the cloze-style problems. It can be seen that the performance of language models was suffered when we simply continue pre-training the models on the commonsense knowledge base. This verifies that the continuing pre-training on the out-of-domain commonsense will cause catastrophic forgetting and injure the effectiveness. On the other hand, using OK-Transformer to introduce commonsense as the extra input significantly improves the accuracy. The results also suggest that ExpBERT is not applicable to large-scale commonsense knowledge bases.

\subsection{Why is OK-Transformer effective?}

We now analyze why OK-Transformer can effectively introduce out-of-domain commonsense without pre-training. We are inspired by an observation of language model fine-tuning LMs~\cite{radiya2020fine}, i.e., the parameters after fine-tuning are close to those before fine-tuning.
Therefore, we argue that the key to effective introduction is whether the parameters of the meta LM is good initialization for the commonsense-enhanced LM, that the parameters do not change much before and after fine-tuning.

\begin{figure}[t]
 \centering
  \includegraphics[scale=.35]{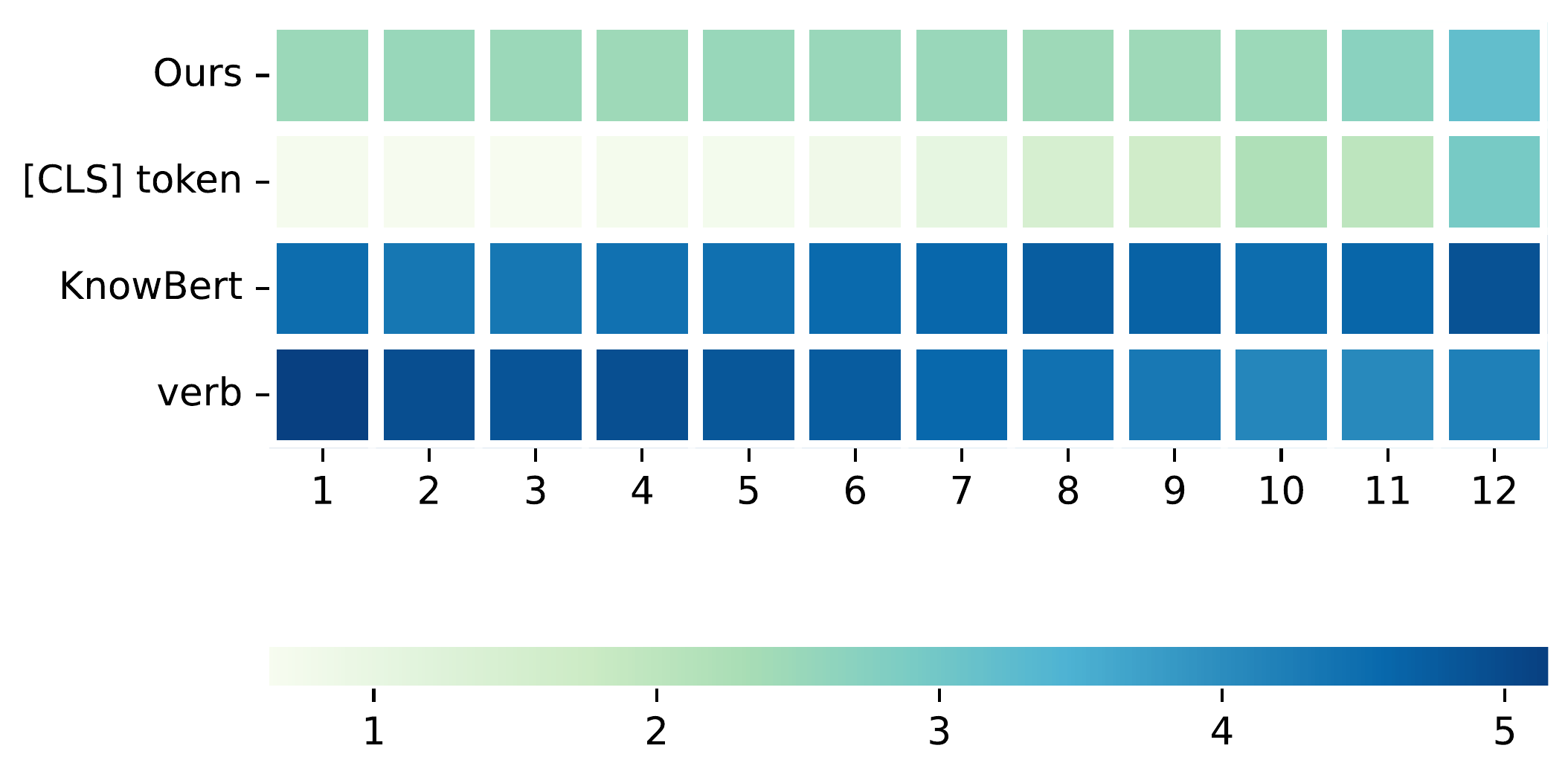}
\caption{$L_1$ distances in parameter space between pre-trained and fine-tuned meta LMs. We show the metrics of $W_I$ across the 12 Transformer layers.}
\label{fig:heatmap}
\end{figure}

\begin{figure}[t]
\centering
\begin{minipage}[t]{0.225\textwidth}
	\centering
  \strut\vspace*{-\baselineskip}\newline\includegraphics[scale=.42]{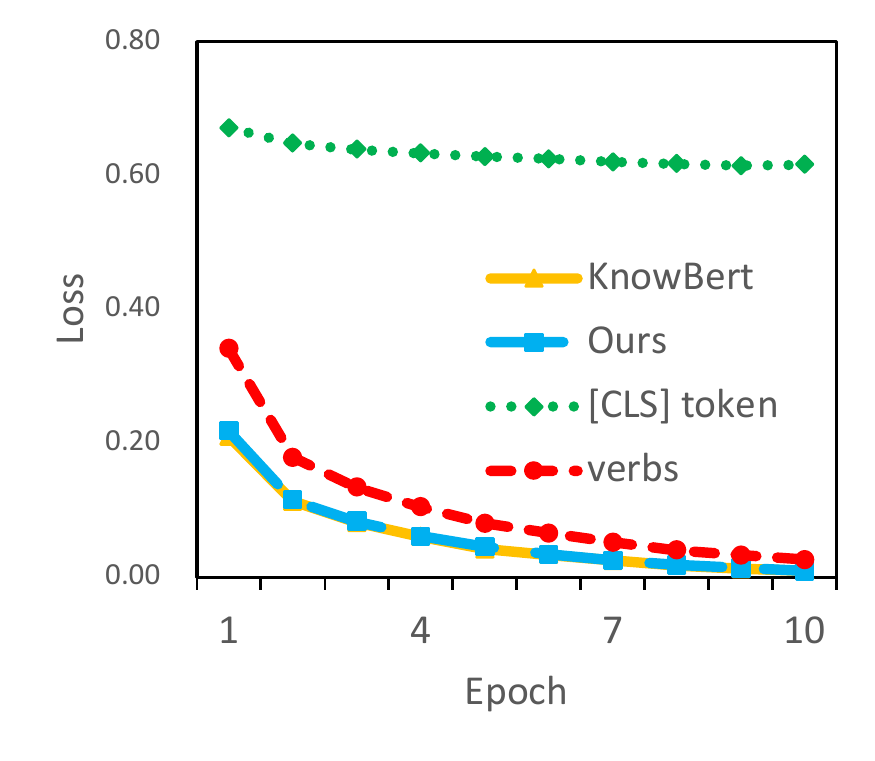}
\caption{Losses of different knowledge integration methods in SST-2. The [CLS] token method does not converge.}
\label{fig:loss}
\end{minipage}
\hspace{0.2cm}
\begin{minipage}[t]{0.225\textwidth}
	\centering
\strut\vspace*{-\baselineskip}\newline\includegraphics[scale=.4]{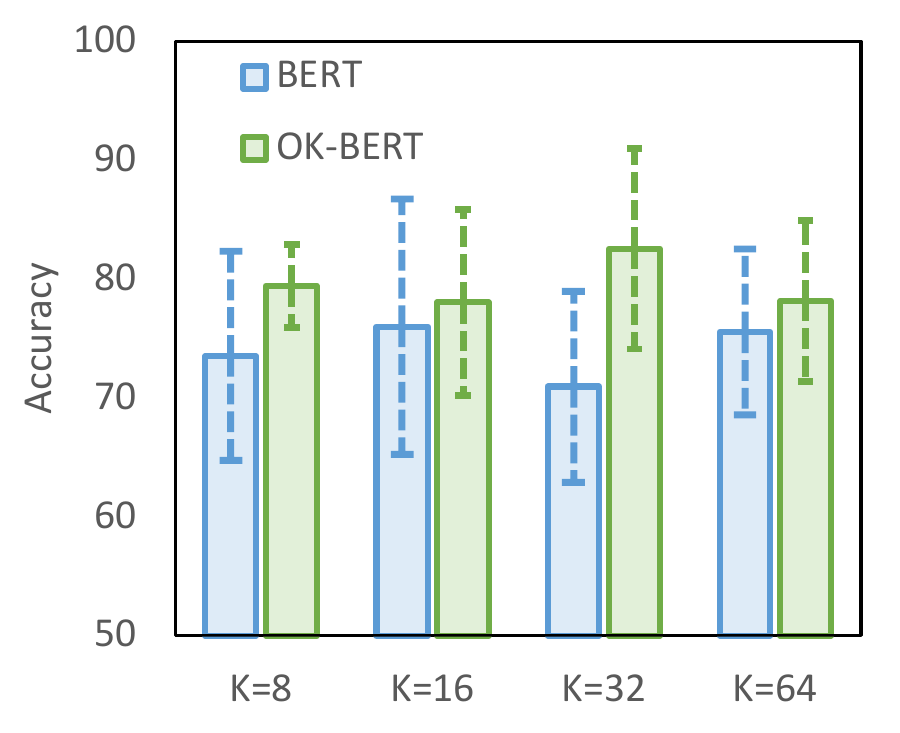}
				\vspace{0.2cm}
\caption{Effect in low-resource commonsense settings with different $k$s over SST-2. }
	\label{fig:low_resource}
\end{minipage}
\end{figure}

To verify this, we compare the parameter changes of different knowledge integration methods. These methods include (1) OK-Transformer, (2) KnowBERT~\cite{peters2019knowledge}, (3) using the original $[CLS]$ token instead of the proposed knowledge token, and (4) abandoning the knowledge token and instead calculating the $cs_{emb}$ of each verb phrase of the target sentence separately, and adding them to these verb phrases' hidden states in $\rm{\bf H}_{i-1}$. We follow~\cite{radiya2020fine} to use the $L1$ as the distance metric. \cite{radiya2020fine} found that the main change in parameters occurs on the $W_I$ matrix of the Transformer. Our experimental results also follow this phenomenon. Therefore, for greater clarity, we only show the distances of the $W_I$ matrices after fine-tune. We show the distances of different methods in Fig.~\ref{fig:heatmap}, and their training losses in Fig.~\ref{fig:loss}.

It can be seen that the distances of OK-Transformer are much smaller than other methods, except the [CLS] token method, which does not converge as shown in Fig.~\ref{fig:loss}. 
This fits our intuition of reducing the parameter variations to introduce external knowledge more effectively.

\subsection{Effect in Low-Resource Commonsense Settings}
Since there is a large number of commonsense descriptions in ATOMIC2020, a large portion of descriptions only occur a few times in the training set. In this subsection, we want to verify for these rare descriptions, can the model still benefit from it? If so, we think it means that the model uses the contextual information of the commonsense to improve the understanding of the commonsense.


To do this, we proposed a low-resource commonsense setting. We evaluate the effect of the model if the training dataset only contains $k=8/16/32/64$ samples. Therefore the frequency of the appeared commonsense descriptions is low. 
In order to exclude the influence of other samples, we only use test samples whose candidate commonsense descriptions have already occurred in the $k$ training samples. For example, when $k=8$, we randomly select $8$ samples from the training set for training, and use all samples in the test set which contains the commonsense of the $8$ training samples for evaluation. We show the results over the SST-2 dataset in Fig.~\ref{fig:low_resource}. It can be seen that our models still benefit from low-frequency commonsense.

\subsection{Does OK-Transformer Provide Interpretability?}

In this subsection, we try to answer if the integration of candidate commonsense descriptions by OK-Transformer is interpretable. To answer this question, we calculate the influence of different commonsense descriptions on the model's predictions. We follow~\cite{wu2020perturbed} to quantify the influence of a commonsense description $c_i$ as: If $c_i$ is removed from $cs(x)$, how much will the prediction change? This change is measured by the Euclidean distance between the prediction by $cs(x)-c_i$ and by $cs(x)$. The greater the change in the prediction, the greater the influence of this commonsense.

\begin{table}[!htb]
\setlength{\tabcolsep}{2.2pt}
\small
\centering
\begin{tabular}{@{}llll@{}}
\toprule
\multicolumn{4}{l}{John \underline{promised} Bill to leave, so an hour later {[}John{]} left.}               \\ \midrule
\multicolumn{4}{l}{PersonX promises PersonY.}   \\
\multicolumn{4}{l}{1. $\cdots$ As a result, PersonX wants to fulfill his promise.} \\ 
\multicolumn{4}{l}{2. $\cdots$ PersonX is seen as truthful}                        \\
\multicolumn{4}{l}{3. $\cdots$ PersonX is seen as trustworthy.}                    \\
\multicolumn{4}{l}{4. $\cdots$ Before, PersonX needed to talk to PersonY.}         \\
\multicolumn{4}{l}{5. $\cdots$ Before, PersonX needed to go to PersonY's house.}   \\ \bottomrule
\end{tabular}
\caption{A case study of top 5 commonsense descriptions.}
\label{tab:interpret}
\end{table}

Through the case studies of the samples in WSC273, we found that although commonsense with higher influence is somewhat interpretable for people, the interpretability is not significant. We show some examples in Table~\ref{tab:interpret}. We believe that this is because some commonsense for people has been learned in pre-training. Therefore, the out-of-domain commonsense that these pre-trained language models need to incorporate for downstream tasks is inconsistent with human understanding.

%% file: ok_conclusion.tex
\section{Conclusion}

In this paper, we study how to use commonsense to enhance the general text representation. We first analyzed the challenges brought by the domain discrepancy of commonsense. Then, we propose OK-Transformer to allow commonsense integration and enhancement. In the experiments, we verified the effectiveness of our proposed models in a variety of scenarios, including commonsense reasoning, general text classification, and low-resource commonsense. Our models consistently outperform the baselines.
We have also empirically analyzed other properties (e.g. interpretability) of the model.

%% file: ok_append.tex


\appendix

\begin{table*}[htb!]
\centering
\begin{tabular}{lccccc}
\toprule[1pt]
Dataset               & WSC     & PDP    & WinoGender & WinoGrande \\ \hline
Dataset size          & 273      & 60     & 720        & 40938/1267 \\
Matched ratio    & 67\%      & 83\%   & 65\%       & 71\%       \\
Average $|cs(x)|$ & 129.71 & 189.68 & 80.63      & 140.56     \\
Average length of $c$ & 17.88   & 17.91  & 16.83      & 17.91   \\
\bottomrule[1pt]
\end{tabular}
\caption{Statistical results on commonsense reasoning datasets.}
\label{tab:stat:wsc}
\end{table*}

\begin{table*}[htb!]
\centering
\begin{tabular}{lcccccc}
\toprule[1pt]
Dataset               & MRPC     & CoLA      & RTE      & QNLI        & STS-B     & SST-2     \\ \hline
Dataset size          & 3668/408 & 8551/1043 & 2490/277 & 104743/5463 & 5749/1500 & 67349/872 \\
Matched ratio    & 59\%     & 40\%      & 72\%     & 52\%        & 56\%      & 25\%      \\
Average $|cs(x)|$ & 80.71    & 84.85     & 122.60   & 81.35       & 117.00    & 83.07     \\
Average length of $c$ & 17.47    & 17.60     & 17.71    & 17.59       & 17.34     & 17.59    \\
\bottomrule[1pt]
\end{tabular}
\caption{Statistical results on sentence classification datasets.}
\label{tab:stat:glue}
\end{table*}

\section{Experimentation Details}

When {\bf continuing pre-training} BERT-continue/RoBERTa-continue in Table~\ref{tab:continue}, we follow~\cite{kocijan19acl} and set learning rate to $1e-5$, batch size to $64$, and train the model for only one epoch.

When {\bf fine-tuning} the models in Sec~\ref{sec:exp:glue} and Sec~\ref{sec:exp:introduction}, we train the models for $10$ epochs. We use grid search to select their learning rates and batch sizes from $\{1e-5, 2e-5, 5e-5\}$ and $\{8, 16, 32,64\}$, respectively. 

\section{Statistics of Commonsense Descriptions}
In Table~\ref{tab:stat:wsc} and Table~\ref{tab:stat:glue}, we report statistics about down-stream tasks and their commonsense descriptions. Our report includes the size of the train/test splits for the downstream tasks, the proportion of samples that matched to at least one commonsense description ({\it Matched proportion}) in each task, the average number of matched commonsense descriptions per sample ({\it Average $|cs(x)|$}), and the average length of each matched commonsense description ({\it Average length of $c$}).

From the results, we found that more than half of the samples matched to at least one commonsense description in most of the datasets. This indicates that the OOD commonsense used in this paper is generalizable to different datasets. Also, the average length of the matched commonsense descriptions is short (about $17$), thus encoding them via Transformer is efficient.

%% file: ok_transformer.bbl
\begin{thebibliography}{35}
\expandafter\ifx\csname natexlab\endcsname\relax\def\natexlab#1{#1}\fi

\bibitem[{Bisk et~al.(2020)Bisk, Zellers, Gao, Choi et~al.}]{bisk2020piqa}
Yonatan Bisk, Rowan Zellers, Jianfeng Gao, Yejin Choi, et~al. 2020.
\newblock Piqa: Reasoning about physical commonsense in natural language.
\newblock In \emph{Proceedings of the AAAI Conference on Artificial
  Intelligence}, volume~34, pages 7432--7439.

\bibitem[{Devlin et~al.(2019)Devlin, Chang, Lee, and
  Toutanova}]{devlin2018bert}
Jacob Devlin, Ming-Wei Chang, Kenton Lee, and Kristina Toutanova. 2019.
\newblock Bert: Pre-training of deep bidirectional transformers for language
  understanding.
\newblock In \emph{NAACL}, pages 4171--4186.

\bibitem[{Emami et~al.(2018)Emami, Trischler, Suleman, and
  Cheung}]{emami2018generalized}
Ali Emami, Adam Trischler, Kaheer Suleman, and Jackie Chi~Kit Cheung. 2018.
\newblock A generalized knowledge hunting framework for the winograd schema
  challenge.
\newblock In \emph{Proceedings of the 2018 Conference of the North American
  Chapter of the Association for Computational Linguistics: Student Research
  Workshop}, pages 25--31.

\bibitem[{Guan et~al.(2020)Guan, Huang, Zhao, Zhu, and
  Huang}]{guan2020knowledge}
Jian Guan, Fei Huang, Zhihao Zhao, Xiaoyan Zhu, and Minlie Huang. 2020.
\newblock A knowledge-enhanced pretraining model for commonsense story
  generation.
\newblock \emph{Transactions of the Association for Computational Linguistics},
  8:93--108.

\bibitem[{Gururangan et~al.(2020)Gururangan, Marasovi{\'c}, Swayamdipta, Lo,
  Beltagy, Downey, and Smith}]{gururangan2020don}
Suchin Gururangan, Ana Marasovi{\'c}, Swabha Swayamdipta, Kyle Lo, Iz~Beltagy,
  Doug Downey, and Noah~A Smith. 2020.
\newblock Don't stop pretraining: Adapt language models to domains and tasks.
\newblock \emph{arXiv preprint arXiv:2004.10964}.

\bibitem[{Guu et~al.(2020)Guu, Lee, Tung, Pasupat, and Chang}]{guu2020realm}
Kelvin Guu, Kenton Lee, Zora Tung, Panupong Pasupat, and Ming-Wei Chang. 2020.
\newblock Realm: Retrieval-augmented language model pre-training.
\newblock \emph{arXiv preprint arXiv:2002.08909}.

\bibitem[{He et~al.(2019)He, Liu, Chen, and Gao}]{he2019hybrid}
Pengcheng He, Xiaodong Liu, Weizhu Chen, and Jianfeng Gao. 2019.
\newblock A hybrid neural network model for commonsense reasoning.
\newblock \emph{arXiv preprint arXiv:1907.11983}.

\bibitem[{Hwang et~al.(2020)Hwang, Bhagavatula, Bras, Da, Sakaguchi, Bosselut,
  and Choi}]{hwang2020comet}
Jena~D Hwang, Chandra Bhagavatula, Ronan~Le Bras, Jeff Da, Keisuke Sakaguchi,
  Antoine Bosselut, and Yejin Choi. 2020.
\newblock Comet-atomic 2020: On symbolic and neural commonsense knowledge
  graphs.
\newblock \emph{arXiv preprint arXiv:2010.05953}.

\bibitem[{Iandola et~al.(2020)Iandola, Shaw, Krishna, and
  Keutzer}]{iandola2020squeezebert}
Forrest Iandola, Albert Shaw, Ravi Krishna, and Kurt Keutzer. 2020.
\newblock Squeezebert: What can computer vision teach nlp about efficient
  neural networks?
\newblock In \emph{Proceedings of SustaiNLP: Workshop on Simple and Efficient
  Natural Language Processing}, pages 124--135.

\bibitem[{Klein and Nabi(2019)}]{klein2019attention}
Tassilo Klein and Moin Nabi. 2019.
\newblock Attention is (not) all you need for commonsense reasoning.
\newblock In \emph{Proceedings of the 57th Annual Meeting of the Association
  for Computational Linguistics}, pages 4831--4836.

\bibitem[{Klein and Nabi(2020)}]{klein2020contrastive}
Tassilo Klein and Moin Nabi. 2020.
\newblock Contrastive self-supervised learning for commonsense reasoning.
\newblock \emph{arXiv preprint arXiv:2005.00669}.

\bibitem[{Kocijan et~al.(2019{\natexlab{a}})Kocijan, Camburu, Cretu, Yordanov,
  Blunsom, and Lukasiewicz}]{kocijan2019wikicrem}
Vid Kocijan, Oana-Maria Camburu, Ana-Maria Cretu, Yordan Yordanov, Phil
  Blunsom, and Thomas Lukasiewicz. 2019{\natexlab{a}}.
\newblock Wikicrem: A large unsupervised corpus for coreference resolution.
\newblock In \emph{Proceedings of the 2019 Conference on Empirical Methods in
  Natural Language Processing and the 9th International Joint Conference on
  Natural Language Processing (EMNLP-IJCNLP)}, pages 4294--4303.

\bibitem[{Kocijan et~al.(2019{\natexlab{b}})Kocijan, Cretu, Camburu, Yordanov,
  and Lukasiewicz}]{kocijan19acl}
Vid Kocijan, Ana-Maria Cretu, Oana-Maria Camburu, Yordan Yordanov, and Thomas
  Lukasiewicz. 2019{\natexlab{b}}.
\newblock A surprisingly robust trick for winograd schema challenge.
\newblock In \emph{The 57th Annual Meeting of the Association for Computational
  Linguistics (ACL)}.

\bibitem[{Levesque et~al.(2012)Levesque, Davis, and
  Morgenstern}]{levesque2012winograd}
Hector Levesque, Ernest Davis, and Leora Morgenstern. 2012.
\newblock The winograd schema challenge.
\newblock In \emph{Thirteenth International Conference on the Principles of
  Knowledge Representation and Reasoning}.

\bibitem[{Liu et~al.(2016)Liu, Jiang, Ling, Zhu, Wei, and
  Hu}]{liu2016commonsense}
Quan Liu, Hui Jiang, Zhen-Hua Ling, Xiaodan Zhu, Si~Wei, and Yu~Hu. 2016.
\newblock Commonsense knowledge enhanced embeddings for solving pronoun
  disambiguation problems in winograd schema challenge.
\newblock \emph{arXiv preprint arXiv:1611.04146}.

\bibitem[{Liu et~al.(2019)Liu, Ott, Goyal, Du, Joshi, Chen, Levy, Lewis,
  Zettlemoyer, and Stoyanov}]{liu2019roberta}
Yinhan Liu, Myle Ott, Naman Goyal, Jingfei Du, Mandar Joshi, Danqi Chen, Omer
  Levy, Mike Lewis, Luke Zettlemoyer, and Veselin Stoyanov. 2019.
\newblock Roberta: A robustly optimized bert pretraining approach.
\newblock \emph{arXiv preprint arXiv:1907.11692}.

\bibitem[{Morgenstern et~al.(2016)Morgenstern, Davis, and
  Ortiz}]{morgenstern2016planning}
Leora Morgenstern, Ernest Davis, and Charles~L Ortiz. 2016.
\newblock Planning, executing, and evaluating the winograd schema challenge.
\newblock \emph{AI Magazine}, 37(1):50--54.

\bibitem[{Murty et~al.(2020)Murty, Koh, and Liang}]{murty2020expbert}
Shikhar Murty, Pang~Wei Koh, and Percy Liang. 2020.
\newblock Expbert: Representation engineering with natural language
  explanations.
\newblock In \emph{Proceedings of the 58th Annual Meeting of the Association
  for Computational Linguistics}, pages 2106--2113.

\bibitem[{Peters et~al.(2019)Peters, Neumann, Logan, Schwartz, Joshi, Singh,
  and Smith}]{peters2019knowledge}
Matthew~E Peters, Mark Neumann, Robert Logan, Roy Schwartz, Vidur Joshi, Sameer
  Singh, and Noah~A Smith. 2019.
\newblock Knowledge enhanced contextual word representations.
\newblock In \emph{Proceedings of the 2019 Conference on Empirical Methods in
  Natural Language Processing and the 9th International Joint Conference on
  Natural Language Processing (EMNLP-IJCNLP)}, pages 43--54.

\bibitem[{Radford et~al.(2019)Radford, Wu, Child, Luan, Amodei, and
  Sutskever}]{radford2019language}
Alec Radford, Jeff Wu, Rewon Child, David Luan, Dario Amodei, and Ilya
  Sutskever. 2019.
\newblock Language models are unsupervised multitask learners.

\bibitem[{Radiya-Dixit and Wang(2020)}]{radiya2020fine}
Evani Radiya-Dixit and Xin Wang. 2020.
\newblock How fine can fine-tuning be? learning efficient language models.
\newblock In \emph{International Conference on Artificial Intelligence and
  Statistics}, pages 2435--2443. PMLR.

\bibitem[{Rudinger et~al.(2018)Rudinger, Naradowsky, Leonard, and
  Van~Durme}]{rudinger2018gender}
Rachel Rudinger, Jason Naradowsky, Brian Leonard, and Benjamin Van~Durme. 2018.
\newblock Gender bias in coreference resolution.
\newblock In \emph{Proceedings of the 2018 Conference of the North American
  Chapter of the Association for Computational Linguistics: Human Language
  Technologies, Volume 2 (Short Papers)}, pages 8--14.

\bibitem[{Sakaguchi et~al.(2019)Sakaguchi, Bras, Bhagavatula, and
  Choi}]{sakaguchi2019winogrande}
Keisuke Sakaguchi, Ronan~Le Bras, Chandra Bhagavatula, and Yejin Choi. 2019.
\newblock Winogrande: An adversarial winograd schema challenge at scale.
\newblock \emph{arXiv preprint arXiv:1907.10641}.

\bibitem[{Sun et~al.(2019)Sun, Qiu, Xu, and Huang}]{sun2019finetune}
Chi Sun, Xipeng Qiu, Yige Xu, and Xuanjing Huang. 2019.
\newblock How to fine-tune bert for text classification?
\newblock In \emph{Proceedings of China National Conference on Computational
  Linguistics}, pages 194--206.

\bibitem[{Talmor et~al.(2019)Talmor, Herzig, Lourie, and
  Berant}]{talmor2019commonsenseqa}
Alon Talmor, Jonathan Herzig, Nicholas Lourie, and Jonathan Berant. 2019.
\newblock Commonsenseqa: A question answering challenge targeting commonsense
  knowledge.
\newblock In \emph{Proceedings of the 2019 Conference of the North American
  Chapter of the Association for Computational Linguistics: Human Language
  Technologies, Volume 1 (Long and Short Papers)}, pages 4149--4158.

\bibitem[{Trinh and Le(2018)}]{trinh2018simple}
Trieu~H Trinh and Quoc~V Le. 2018.
\newblock A simple method for commonsense reasoning.
\newblock \emph{arXiv preprint arXiv:1806.02847}.

\bibitem[{Vaswani et~al.(2017)Vaswani, Shazeer, Parmar, Uszkoreit, Jones,
  Gomez, Kaiser, and Polosukhin}]{vaswani2017attention}
Ashish Vaswani, Noam Shazeer, Niki Parmar, Jakob Uszkoreit, Llion Jones,
  Aidan~N Gomez, {\L}ukasz Kaiser, and Illia Polosukhin. 2017.
\newblock Attention is all you need.
\newblock In \emph{Proceedings of the 31st International Conference on Neural
  Information Processing Systems}, pages 6000--6010.

\bibitem[{Wang et~al.(2018)Wang, Singh, Michael, Hill, Levy, and
  Bowman}]{wang2018glue}
Alex Wang, Amanpreet Singh, Julian Michael, Felix Hill, Omer Levy, and Samuel
  Bowman. 2018.
\newblock Glue: A multi-task benchmark and analysis platform for natural
  language understanding.
\newblock In \emph{Proceedings of the 2018 EMNLP Workshop BlackboxNLP:
  Analyzing and Interpreting Neural Networks for NLP}, pages 353--355.

\bibitem[{Wang et~al.(2020)Wang, Fang, Sun, Gan, Cheng, Liu, and
  Jiang}]{wang2020cross}
Shuohang Wang, Yuwei Fang, Siqi Sun, Zhe Gan, Yu~Cheng, Jingjing Liu, and Jing
  Jiang. 2020.
\newblock Cross-thought for sentence encoder pre-training.
\newblock In \emph{Proceedings of the 2020 Conference on Empirical Methods in
  Natural Language Processing (EMNLP)}, pages 412--421.

\bibitem[{Wang et~al.(2019)Wang, Zhang, Shen, Liu, Liu, Gao, and
  Jiang}]{wang2019unsupervised}
Shuohang Wang, Sheng Zhang, Yelong Shen, Xiaodong Liu, Jingjing Liu, Jianfeng
  Gao, and Jing Jiang. 2019.
\newblock Unsupervised deep structured semantic models for commonsense
  reasoning.
\newblock \emph{arXiv preprint arXiv:1904.01938}.

\bibitem[{Wolf et~al.(2020)Wolf, Chaumond, Debut, Sanh, Delangue, Moi, Cistac,
  Funtowicz, Davison, Shleifer et~al.}]{wolf2020transformers}
Thomas Wolf, Julien Chaumond, Lysandre Debut, Victor Sanh, Clement Delangue,
  Anthony Moi, Pierric Cistac, Morgan Funtowicz, Joe Davison, Sam Shleifer,
  et~al. 2020.
\newblock Transformers: State-of-the-art natural language processing.
\newblock In \emph{Proceedings of the 2020 Conference on Empirical Methods in
  Natural Language Processing: System Demonstrations}, pages 38--45.

\bibitem[{Wu et~al.(2020)Wu, Chen, Kao, and Liu}]{wu2020perturbed}
Zhiyong Wu, Yun Chen, Ben Kao, and Qun Liu. 2020.
\newblock Perturbed masking: Parameter-free probing for analyzing and
  interpreting bert.
\newblock In \emph{Proceedings of the 58th Annual Meeting of the Association
  for Computational Linguistics}, pages 4166--4176.

\bibitem[{Xiong et~al.(2019)Xiong, Du, Wang, and
  Stoyanov}]{xiong2019pretrained}
Wenhan Xiong, Jingfei Du, William~Yang Wang, and Veselin Stoyanov. 2019.
\newblock Pretrained encyclopedia: Weakly supervised knowledge-pretrained
  language model.
\newblock In \emph{International Conference on Learning Representations}.

\bibitem[{Zhang et~al.(2019)Zhang, Han, Liu, Jiang, Sun, and
  Liu}]{zhang2019ernie}
Zhengyan Zhang, Xu~Han, Zhiyuan Liu, Xin Jiang, Maosong Sun, and Qun Liu. 2019.
\newblock Ernie: Enhanced language representation with informative entities.
\newblock In \emph{Proceedings of the 57th Annual Meeting of the Association
  for Computational Linguistics}, pages 1441--1451.

\bibitem[{Zhou et~al.(2018)Zhou, Young, Huang, Zhao, Xu, and
  Zhu}]{zhou2018commonsense}
Hao Zhou, Tom Young, Minlie Huang, Haizhou Zhao, Jingfang Xu, and Xiaoyan Zhu.
  2018.
\newblock Commonsense knowledge aware conversation generation with graph
  attention.
\newblock In \emph{IJCAI}, pages 4623--4629.

\end{thebibliography}
